\documentclass{article} 
\usepackage{iclr2026_conference,times}

\usepackage{amsmath,amsfonts,bm}

\def\eqref#1{equation~\ref{#1}}

\def\1{\bm{1}}

\DeclareMathAlphabet{\mathsfit}{\encodingdefault}{\sfdefault}{m}{sl}
\SetMathAlphabet{\mathsfit}{bold}{\encodingdefault}{\sfdefault}{bx}{n}

\newcommand{\softmax}{\mathrm{softmax}}

\usepackage{hyperref}
\usepackage{url}
\usepackage{multirow}
\usepackage{multicol}
\usepackage{booktabs} 
\usepackage{graphicx}
\usepackage{subcaption}

\title{Draft, Verify, \& Improve \\Toward Training-Aware Speculative Decoding }

\author{Shrenik Bhansali, Larry Heck \\
Georgia Institute of Technology\\
\texttt{\{sbhansali8,larryheck\}@gatech.edu} \\
}

\iclrfinalcopy
\begin{document}

\maketitle

\begin{abstract}

Autoregressive (AR) decoding is a major latency bottleneck for large language models. Speculative decoding (SD) accelerates AR by letting a drafter propose multi-token blocks that a verifier accepts or rejects. However, many SD systems require heavy offline training or extra components. These choices raise data/compute cost and can yield brittle drafters under distribution drift.
We introduce \emph{Draft, Verify, \& Improve (DVI)}, a training-aware self-speculative framework that combines inference with continual online learning. We partition an LLM into a drafter and a verifier, and during generation, verifier accept/reject decisions are converted into supervision signals and used to update the drafter head. 
A simple \emph{KL$\rightarrow$RL} schedule bootstraps calibration via online distillation and then adds reward-masked cross-entropy with a on-policy policy-gradient term, preserving lossless, single model deployment.
On Spec-Bench, DVI achieves a $2.16\times$ wall-time speedup, on par with SoTA approaches like EAGLE-2, while orders of magnitude less data for training, and ablations show that DVI outperforms KL-only online distillation. 
DVI demonstrates that \emph{training-aware} self-speculation can deliver state-of-the-art, lossless speedups with minimal training overhead. 

Code will be released upon publication.





\end{abstract}

\section{Introduction}

Large Language Models (LLMs) are continuing to rise in popularity, finding new use-cases ranging from powering conversational assistants to web agents. However, models continue to suffer high latency due to their autoregressive (AR) nature.  AR decoding acts as the primary bottleneck for LLM inference as each token depends on the previous one, causing end-to-end latency to scale sequence length and per-step compute.

Speculative decoding (SD) addresses this by proposing multiple tokens at once instead of one token at a time \citep{chen2023accelerating, leviathan2023fast}. SD uses  a lightweight drafting model, and uses the larger, high-fidelity target model to verify or reject each drafted token. The accepted tokens are then committed and generated, preserving target-model correctness while reducing wall-clock time. While SD has exploded in popularity, many SD works still face practical limitations. Speedups are dependent on high token acceptance rates, which in turn hinge on the drafter’s fidelity. If the drafted tokens diverge, the verifier takes over, and falls back to AR decoding, plus the added drafting computation. 

In traditional SD, many works train a dedicated drafting model, or distill a drafting model from the larger target model \citep{chen2023accelerating, leviathan2023fast}. Other works explore self-speculation, where the model itself is partitioned into drafting layers and verification layers \citep{zhang2023draft, liu2024kangaroolosslessselfspeculativedecoding}. Regardless of which approach, most techniques rely heavily on a robust training phase to train the drafting model. This heavy offline drafter training introduces practical gaps for SD. 

Many SD pipelines require training a drafting model for multiple epochs over large datasets, or intensive model distillation from full teacher distributions over data  \citep{ankner2024hydra, cai2024medusa, li2024eagle, li2024eagle-2, zhang2023draft}. Even training a lightweight adapter requires hours of training \citep{liu2024kangaroolosslessselfspeculativedecoding}. 

This reliance on heavy offline training enables drafter brittleness under distribution shift. As conversation, task, or traffic drifts, the fixed drafters lose acceptance, killing any potential speedup, if the distribution is not covered by the training data \citep{liu2024onlinespeculativedecoding}. These slowdowns compound in traditional SD settings, where the pipeline relies on separate drafter and verifier models, due to the added computational cost of running both models, and the ensuring KV cache overhead, system complexity, and increased memory demands.

Therefore, we propose Draft, Verify, \& Improve (DVI), a training-aware self-speculative decoding framework. DVI splits a single backbone model at an intermediate layer, attaching a drafter head and a frozen verifier head. This partitioning creates a computationally efficient drafting model and verifying model. At inference, the drafter proposes multi-token blocks which the verifier accepts or rejects. 

Whereas most SD pipelines will just commit the accepted tokens and move on, DVI instead treats these commit decisions as learning signals: the drafter is updated online from accept/reject feedback while the verifier remains unchanged. This preserves a one-model serving geometry with no auxiliary drafter or extra KV cache while enabling continual drafter adaptation to live traffic.

During generation, the drafter proposes \(k_{\text{spec}}\) tokens from prefix tokens \(h_k\); the frozen verifier evaluates the same prefix from \(h_L\) and commits the longest agreeing prefix between the drafter and verifier. The frozen verifier ensures speculation remains lossless. 
DVI logs the per-token state at \(h_k\), denoting a accept or reject, into a lightweight buffer and performs small, frequent updates to the drafter using a \emph{KL\(\rightarrow\)RL schedule}: a KL-guided warmup to the verifier distribution followed by reward-masked cross-entropy and an on-policy policy-gradient term \cite{williams1992simple}.
Training can be done in an online manner, as it mirrors the inference workflow of drafting and verifying, minimizing train/serve skew.

We evaluate DVI on Spec-Bench, a public SD benchmark suite spanning translation, summarization, QA, math, and retrieval-augmented generation. DVI achieves near-\(2\times\) average wall-time speedups with a single pass over a small prompt stream.
Acceptance improves steadily during online updates, and the method maintains strong performance without any additional offline training or auxiliary models.

We perform ablations to verify that our training pipeline is robust. We demonstrate that optimizing the drafter purely by distillation or purely by sparse rewards is insufficient. A KL/distillation-only variant, performing online KD from the frozen verifier, creates a speedup, but falls short of matching DVI performance. 

These results motivate DVI’s \emph{KL$\rightarrow$RL} schedule: an online KD warmup that calibrates the shallow drafter in the verifier’s logit space, followed by a on-policy correction that assigns credit only where speculation succeeds. This combination overcomes both the KL plateau and the instability of sparse-reward training, directly optimizing acceptance where it matters.

\paragraph{Contributions:}
\begin{itemize}
  \item We propose DVI, a self-speculation method with a frozen verifier and an online-learned drafter head that converts commit decisions into self-supervision.
  \item DVI offers a data-efficient, cheap method to train a SD model. We create competitive speedups using a small number of live prompts, with no separate offline dataset or long pretraining.
  \item We validate DVI's effectiveness by performing experiments with Spec-Bench, demonstrating competitive speedups compared to other SoTA SD methods.  
\end{itemize}

The paper is structured as follows:
Section~\ref{sec:background} introduces related works.
Section~\ref{sec:method} presents the DVI pipeline, objective, and training schedule.
Section~\ref{experiments} reports speedups, comparisons against other SD methods, and ablations.
Section~\ref{conclusion} concludes the paper.

\section{Related Work} \label{sec:background}

\subsection{Speculative Decoding Basics}

 Traditional modern speculative decoding (SD) accelerates autoregressive generation by letting a small draft model propose a block of $k$ tokens and a large target model verify them in parallel.
 The target model then commits the longest agreeing prefix and continues generation. In this setting,  the committed tokens from the verifier are exactly from the target model’s decoding distribution (e.g., greedy or sampling) — creating “lossless” speculation \citep{leviathan2023fast, chen2023accelerating}.

\citet{liu2024onlinespeculativedecoding},  periodically fine-tunes an external drafter model online via knowledge distillation (KD) \citep{liu2024onlinespeculativedecoding}. By aligning to live query distributions, it shows that continual adaptation is possible, albeit with an auxiliary drafter and pretraining warm-start.
 
\subsection{Self-Speculation}
Instead of a separate drafting model, self-speculation splits a single backbone model into shallow drafting layers, and deep verification layers. 
\citet{zhang2023draft} introduced self-speculation for LLMs by skipping or shortening intermediate layers to form a fast internal drafter and then running the full model to verify. 

This simpler approach removes the complexity of managing multiple models, finding suitable drafting models for target models, and reduces computation. 

Later work improves upon this, with more complex and flexible architectures.
For example, \citet{liu2024kangaroolosslessselfspeculativedecoding} achieves self speculation by adding a lightweight adapter over the shallow subnetwork and introducing dynamic early-exit during drafting to curb drafter latency when confidence is low.

\subsection{Other Acceleration Families}

Instead of speculating, \citet{cai2024medusa} augments the backbone model with multiple time-independent heads that predict multi-step continuations. The resulting branches are verified with  tree-attention verifier every step.
\citet{ankner2024hydra} replaces Medusa’s independent heads with sequentially dependent heads, improving draft accuracy (and thus acceptance) under the same verification framework.

\citet{li2024eagle} drafts features (second-to-top-layer states) one step ahead and uses the target LM head to form tokens, producing well-calibrated draft trees. \citet{li2024eagle-2} adds context-aware dynamic trees, yielding improved speedups. While these approaches create massive speedups, they also require significant amounts of training, even relative to approaches like Medusa.

\section{DVI: Draft $\rightarrow$ Verify $\rightarrow$ Improve}
\label{sec:method}

\begin{figure}[t]
\begin{center}
\includegraphics[width=\textwidth]{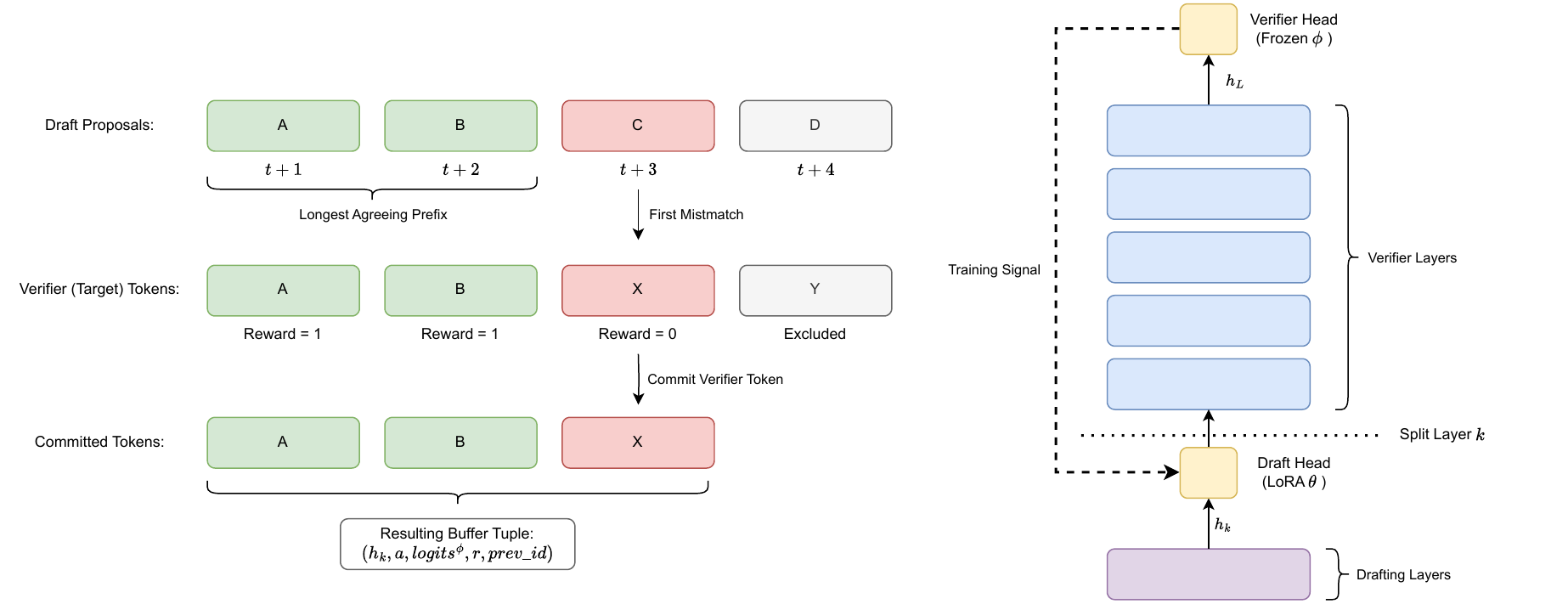}
\end{center}
\caption{
\emph{\textbf{Left:}} Multi-token speculation, where the drafter proposes a block of tokens and the verifier accepts the longest agreeing prefix before emitting the first mismatch. We log one tuple per drafted position up to and including the first reject, $(h_k, a, \text{logits}^{\phi}, r, \text{prev\_id})$, with $r{=}1$ for accepted tokens and $r{=}0$ for the first reject. This converts verifier feedback into continual self-supervision. 
\emph{\textbf{Right:}} DVI architecture, where the backbone is split at layer $k$, with shallow drafting layers (purple) feeding the LoRA draft head $p_\theta(\cdot\mid h_k)$ and deep verification layers (blue) feeding the frozen verifier head $p_\phi(\cdot\mid h_L)$. 
The logged tuples from the rollout buffer drive updates to the draft head, while the verifier and backbone remain fixed. This closes the loop between online speculation and training, ensuring adaptation without additional models or offline data.}
\end{figure}

This section formalizes \emph{Draft, Verify, \& Improve (DVI)} for self-speculative decoding using a single backbone model with a LoRA-parameterized draft head.

\subsection{Preliminaries and Notation}
\label{sec:prelims}

Consider a decoder-only language model with transformer layers indexed $0{:}L$. We choose a split index $k$ with $0<k<L$ and write the shallow and deep hidden states as
\[
h_{k,t} = f_{0\to k}(x_{0:t}), \qquad
h_{L,t} = f_{k\to L}(h_{k,t}),
\]
where $x_{0:t}$ is the token prefix, $f_{0\to k}$ is the \emph{draft path} (layers $0{\to}k$), and $f_{k\to L}$ is the \emph{target path} (layers $k{\to}L$). 

We attach two vocabulary classifiers:
\[
p_\phi(\cdot \mid h_{L,t}) = \softmax(W^{(V)} h_{L,t}) \quad \text{(target path; \emph{frozen})},
\]
\[
p_\theta(\cdot \mid h_{k,t}) = \softmax\Big((W^{(S)} + \gamma_s A_s B_s)\,h_{k,t}\Big) \quad \text{(draft path; \emph{trainable LoRA})}.
\]

Here $W^{(S)}$ is a frozen base projection at the draft path output, and $(A_s,B_s)$ are trainable LoRA modules \citep{hu2021loralowrankadaptationlarge}.
Throughout training, \emph{only} $\theta\equiv\{A_s,B_s\}$ is updated online; all backbone weights remain fixed. LoRA is applied only to the draft head at $h_{k,t}$.

We adopt the canonical longest-prefix verification used in SD: at each position $t$, the target path deterministically emits the next token under a fixed sampler. Like many other SD works,we employ \emph{greedy} decoding,
\[
y^\star_{t+1} \,=\, \arg\max_{y} p_\phi(y\mid h_{L,t}),
\]
and the procedure is lossless because verification preserves the target sampler’s output. Additionally, we consider only \emph{single-sequence} verification (no tree search). 

\subsection{Self-Speculative Factorization}
\label{sec:factorization}

As mentioned before, DVI is self-speculative: the backbone model is partitioned into a draft path (layers $0{\to}k$) and a target path (layers $k{\to}L$). 
The LoRA-augmented draft head at $h_{k,t}$ proposes tokens quickly; the frozen target head at $h_{L,t}$ verifies the tokens. This single-model setup avoids separate drafting and target models, the ensuing extra KV cache, and larger system level complexity, while providing a LoRA adapters for online learning.

Drafting a $k_{\text{spec}}$-token block requires one shallow forward; if $m$ tokens are accepted, the deep computation is amortized over those $m$ outputs in a single verification pass.

\subsection{Speculative Rollout and Learning Signal}
\label{sec:rollout}
We can see what one speculation and verification looks like given the notation above. 

At decoding step $t$, we compute the shallow state once given the prefix tokens (commonly user input or training data),
\[
h_{k,t} = f_{0\to k}(x_{0:t}).
\]
The draft path rolls out up to $k_{\text{spec}}$ candidates
\[
\tilde t_{t+1:t+k_{\text{spec}}} \sim p_\theta(\cdot\mid h_{k,\cdot}),
\]
 
Given the drafted candidates, we verify for each prefix length $i=1,\dots,k_{\text{spec}}$,
\[
h_{L,t+i-1} = f_{k\to L}(h_{k,t+i-1}), \qquad y^\star_{t+i}=\arg\max p_\phi(\cdot\mid h_{L,t+i-1}).
\]
Let
\[
m \,=\, \max\Big\{i\in\{0,\dots,k_{\text{spec}}\}: \tilde t_{t+j}=y^\star_{t+j}\ \text{for all } j\le i\Big\}.
\]
We the commit the $m$ agreeing tokens and reject everything at and after the first mismatch by emitting $y^\star_{t+m+1}$. Decoding then continues autoregressively from the new prefix.

The resulting accept/reject outcomes yield a clean, low-variance signal. For drafted positions up to and including the first reject, we log to an \emph{online replay buffer}
\[
\big(h_{k,t+i-1},\ a_{t+i}=\tilde t_{t+i},\ \text{logits}^\phi_{t+i},\ r_{t+i},\ i\big),
\]
where $\text{logits}^\phi_{t+i}$ are target-path logits and
\[
r_{t+i} =
\begin{cases}
1, & 1\le i\le m \quad \text{(accepted)},\\
0, & i=m+1 \quad \text{(first reject)},\\
\text{undefined}, & i>m+1 \ \ \text{(counterfactual; not verified)}.
\end{cases}
\]
We exclude $i>m{+}1$ from supervised terms to avoid counterfactual bias. The buffer mirrors inference (same $k_{\text{spec}}$ and commit rule), reducing train/serve skew, and generalizing to online learning. 

\subsection{Objectives and Update Schedule}
\label{sec:objectives}

To prevent an RL-cold start with low acceptance, early updates imitate the target path (online KD) to stabilize gradients in the low-rank subspace; later, the draft path optimizes acceptance on observed traffic. We implement a composite objective with a KL-to-RL schedule acting on the LoRA adapters $\theta$.

For a minibatch $\mathcal{B}$ sampled from the online buffer, we minimize
\[
\mathcal{L}_{\text{fast}}
= \lambda_{\text{pg}}\,\mathcal{L}_{\text{pg}}
+ \lambda_{\text{kl}}\,\mathrm{KL}\!\big(p_\theta \,\|\, p_\phi^{(\tau)}\big)
+ w_{\text{ce}}\,\mathcal{L}_{\text{CE}}
- w_{\text{ent}}\,\mathcal{H}[p_\theta],
\]
where $p_\theta(\cdot\mid h_{k,t})=\softmax\!\big((W^{(S)}+\gamma_sA_sB_s)h_{k,t}\big)$ and $p_\phi^{(\tau)}=\text{softmax}(\text{logits}^\phi/\tau)$. 

The reward-masked term
\[
\mathcal{L}_{\text{pg}} \,=\, -\frac{1}{|\mathcal{P}|}\sum_{i\in\mathcal{P}}\log p_\theta(a_{t+i}\mid h_{k,t+i-1})
\]
uses only \emph{accepted} positions $\mathcal{P}$ to focus credit where speculation succeeded, and $\mathrm{KL}(p_\theta\|p_\phi^{(\tau)})$ performs online KD for calibration.

Once the cold start is avoided, we add a light, \emph{on-policy} correction on fresh tuples:
\[
\mathcal{L}_{\text{policy}}
= w_{\text{rl}}\,\mathbb{E}_{(s,a,r)\sim \text{on-policy}}\!\big[-(r-b)\,\log p_\theta(a\mid s)\big]
+ \beta(t)\,\mathrm{KL}\!\big(p_\theta \,\|\, p_\phi\big),
\]
where $b$ is a baseline for variance reduction (we use an EMA of recent rewards) and $\beta(t)$ gently decays to retain calibration. We include both accepted and first-reject positions; positions beyond the first reject are excluded (counterfactual).

We anneal the relative weights over wall-clock updates $t$:
\[
(\lambda_{\text{pg}},\lambda_{\text{kl}})(t)=
\begin{cases}
(0,\ \lambda_0), & t<T_{\text{warmup}},\\
\Big(\!\frac{t-T_{\text{warmup}}}{T_{\text{ramp}}}\lambda_{\text{pg}}^{\max},\ 
\lambda_0 - \frac{t-T_{\text{warmup}}}{T_{\text{ramp}}}(\lambda_0-\lambda_{\text{kl}}^{\min})\!\Big), & \text{ramp},\\
(\lambda_{\text{pg}}^{\max},\ \lambda_{\text{kl}}^{\min}), & \text{after}.
\end{cases}
\]
Warmup emphasizes online KD to avoid unstable gradients in a misaligned subspace; the ramp increases reward-masked learning that directly raises acceptance.

\section{Experiments}\label{experiments}

\subsection{Experimental Setup}

All experiments use the Spec-Bench benchmark \citep{xia-etal-2024-unlocking}. Spec-Bench is a public speculative decoding benchmark that measures model speedups and accepted tokens over six evaluation settings (MT-Bench, Translation, Summarization, QA, Math, and RAG). All experiments use the same base Vicuna-7B~\citep{zheng2023judgingllmasajudgemtbenchchatbot} model as requested by the benchmark, with the serving configuration fixed across all methods (identical tokenizers, decoding policies, context limits, and so on).

\begin{table}[h]
\caption{Comparison of training data budgets across speculative decoding methods. 
DVI uses orders of magnitude fewer prompt exposures than prior approaches.}
\label{tab:training-data}
\begin{center}
\begin{tabular}{lrrrrr}
\multicolumn{1}{l}{  Method} &
\multicolumn{1}{c}{  ShareGPT} &
\multicolumn{1}{c}{  Epochs} &
\multicolumn{1}{c}{  Prompt } &
\multicolumn{1}{c}{  Optimiser } &
\multicolumn{1}{c}{  Relative } \\
 &
\multicolumn{1}{c}{  Samples} &
 &
\multicolumn{1}{c}{  exposures} &
\multicolumn{1}{c}{   steps} &
\multicolumn{1}{c}{   budget} \\
\hline \\[-0.9em]
DVI (our work) & 2{,}000 & 1  & 2{,}000     & 2{,}000     & 1$\times$ \\
Medusa~\citep{cai2024medusa}          & 60{,}000 & 2  & 120{,}000   & $\approx$945    & $\sim$60$\times$ more \\
Kangaroo~\citep{liu2024kangaroolosslessselfspeculativedecoding}        & 60{,}000  & 20 & 1{,}200{,}000 & $\approx$4{,}700  & $\sim$600$\times$ more \\
EAGLE~\citep{li2024eagle}           & 60{,}000  & 40 & 2{,}400{,}000 & $\approx$300{,}000 & $\sim$1{,}200$\times$ more \\
\end{tabular}
\end{center}
\end{table}

For the DVI model, we split the model such that layer 2 is the drafter, while layers 3 through 32 are the verifier. Since the verifier is the frozen baseline weights, we can guarantee lossless speculation. In the experimental setting, we use a drafting proposal depth of 4 ($k_{\text{spec}}{=}4$). We train the DVI model for 2000 steps over 2000 prompts, such that the model sees each prompt only once.

\begin{table}[b]
  \centering
  \caption{Spec-Bench comparison across speculative decoding methods. Each cell shows mean accepted tokens (MAT) and walltime speedup; the rightmost column reports the overall average speedup.}
  \label{tab:specbench-results}
  \vspace{2pt}
  \setlength{\tabcolsep}{4pt}
  \renewcommand{\arraystretch}{1.05}
  \makebox[\linewidth][c]{%
    \resizebox{\linewidth}{!}{%
      \begin{tabular}{lccccccccccccc}
        \toprule
        \multirow{2}{*}{Method} &
        \multicolumn{2}{c}{MT Bench} &
        \multicolumn{2}{c}{Translation} &
        \multicolumn{2}{c}{Summarization} &
        \multicolumn{2}{c}{QA} &
        \multicolumn{2}{c}{Math} &
        \multicolumn{2}{c}{RAG} &
        \multirow{2}{*}{Avg.} \\
        \cmidrule(lr){2-3} \cmidrule(lr){4-5} \cmidrule(lr){6-7}
        \cmidrule(lr){8-9} \cmidrule(lr){10-11} \cmidrule(lr){12-13}
        & MAT & Speedup & MAT & Speedup & MAT & Speedup & MAT & Speedup & MAT & Speedup & MAT & Speedup & \\
        \midrule
        EAGLE-2 & 4.75 & \textbf{2.64$\times$} & 3.22 & 1.73$\times$ & 3.96 & \textbf{2.15$\times$} & 3.70 & 1.96$\times$ & 4.73 & \textbf{2.59$\times$} & 4.09 & 2.02$\times$ & \textbf{2.18$\times$} \\
        EAGLE-1 & 3.83 & 2.38$\times$ & 2.84 & 1.72$\times$ & 3.32 & 2.03$\times$ & 3.12 & 1.88$\times$ & 3.87 & 2.39$\times$ & 3.27 & 1.88$\times$ & 2.05$\times$ \\
        Hydra   & 3.59 & 2.31$\times$ & 2.80 & 1.80$\times$ & 2.70 & 1.73$\times$ & 2.85 & 1.84$\times$ & 3.61 & 2.31$\times$ & 2.90 & 1.74$\times$ & 1.96$\times$ \\
        Medusa  & 2.51 & 1.91$\times$ & 2.13 & 1.59$\times$ & 2.02 & 1.53$\times$ & 2.09 & 1.59$\times$ & 2.50 & 1.88$\times$ & 2.10 & 1.48$\times$ & 1.66$\times$ \\
        PLD     & 1.69 & 1.61$\times$ & 1.10 & 1.03$\times$ & 2.72 & 2.54$\times$ & 1.37 & 1.14$\times$ & 1.86 & 1.59$\times$ & 1.72 & 1.85$\times$ & 1.62$\times$ \\
        SpS     & 2.33 & 1.62$\times$ & 1.46 & 1.10$\times$ & 2.44 & 1.65$\times$ & 2.17 & 1.45$\times$ & 2.20 & 1.46$\times$ & 2.31 & 1.63$\times$ & 1.48$\times$ \\
        \midrule
        DVI (ours) & 3.07 & 1.97$\times$ & 3.53 & \textbf{2.24$\times$} & 3.55 & 2.02$\times$ & 3.61 & \textbf{2.14$\times$} & 3.04 & 2.02$\times$ & 3.53 & \textbf{2.58$\times$} & 2.16$\times$ \\
        \bottomrule
      \end{tabular}%
    }
  }
\end{table}

Competing methods are taken \emph{as-is} from their public checkpoints and implementations surfaced through Spec-Bench and Hugging Face. We do not retrain baselines; where the harness exposes a method knob (e.g., draft depth), we use the recommended defaults. 

We train DVI on 2{,}000 samples of ShareGPT, aligning our training data with other methods in the literature. Methods like EAGLE, Medusa, Hydra, and Kangaroo all train their models on ShareGPT as well. We illustrate in Table \ref{tab:training-data} that the competing methods are trained for orders of magnitude longer than DVI, in an offline setting. Comparatively, DVI sees a fraction of the prompts that the other methods do, making inherently cheaper to train and more data efficient than other methods.

Following Spec-Bench, we report mean accepted tokens (MAT) per verification step, and wall-time speedup versus the baseline model’s standard autoregressive decoding. Note that MAT is not a one-to-one predictor of model speedup. Having a deeper drafting model for example, will yield a higher MAT, but at the cost of added computation, risking a lower speedup. Methods that propose more tokens will inherently have higher MAT, but it may also induce extra computation that offsets this advantage. 

More implementation details are noted in Appendix~\ref{app:setup}.

\subsection{Results}

DVI achieves competitive walltime speedups with methods like EAGLE-2, while remaining lossless, and using orders-of-magnitude less training data. As shown in Table~\ref{tab:specbench-results}, DVI attains a highest speedup in Translation, QA, and RAG, while EAGLE-2 leads in MT Bench, Summarization, and Math. DVI's average speedup of 2.16$\times$ is on par with EAGLE-2’s 2.18$\times$ despite EAGLE-2 having a $1000\times$ larger training budget as shown in Table~\ref{tab:training-data}.

At a more granular level, these results indicate that DVI is particularly effective for workloads with strong local lexical structure and retrieval grounding (assistant-style tasks), whereas deep tree drafting is more beneficial for long-range reasoning and structured planning. 

We should not discount the role that training data plays. ShareGPT covers a wide range and variety of tasks, and the 2{,}000 samples we take from it are not necessarily evenly distributed. Therefore, DVI may perform better on tasks like MT Bench if exposed to more multi-turn conversations and so on.

Although EAGLE-2 attains a higher mean accepted tokens (MAT), its average wall-time speedup is only marginally higher than DVI. This gap implies that DVI converts agreement into throughput more efficiently. With a shallow drafter and small proposal depth, DVI avoids the extra tree-building and verification work required by multi-head or tree-based approaches, yielding a higher speedup-per-accepted-token and explaining why similar end-to-end gains emerge despite lower MAT.

\subsection{Ablations}
\label{sec:ablations}

We isolate the contribution of each training signal in DVI by ablating the objective into three single-term variants: 
\begin{enumerate}
    \item KL-only, equivalent to online distillation.
    \item PG-only, which is simply on-policy REINFORCE.
    \item CE-only, reward-masked cross entropy.
\end{enumerate}

For these ablations, all runs share the exact same backbone, split layer $k$, proposal depth $k_{\text{spec}}{=}4$, optimizer, batch size, data stream, and hardware as our main setup.

We first observe the learning dynamics of the model, captured by the batch acceptance rate throughout training. We then benchmark the resulting models on Spec-Bench to get their end-state MAT and wall-time speedup.
We define the batch acceptance rate to be the fraction of drafted tokens accepted in each optimization step (higher is better). Note that batch acceptance rate is not a perfect indicator model performance on Spec-Bench, as the speculative batches are directly from training on ShareGPT. Thus, models can attain higher batch acceptance without becoming robust on other distributions, like tasks in Spec-Bench. 

\begin{figure*}[h]
  \centering
  \begin{subfigure}[h]{0.32\textwidth}
    \centering
    \includegraphics[width=\linewidth]{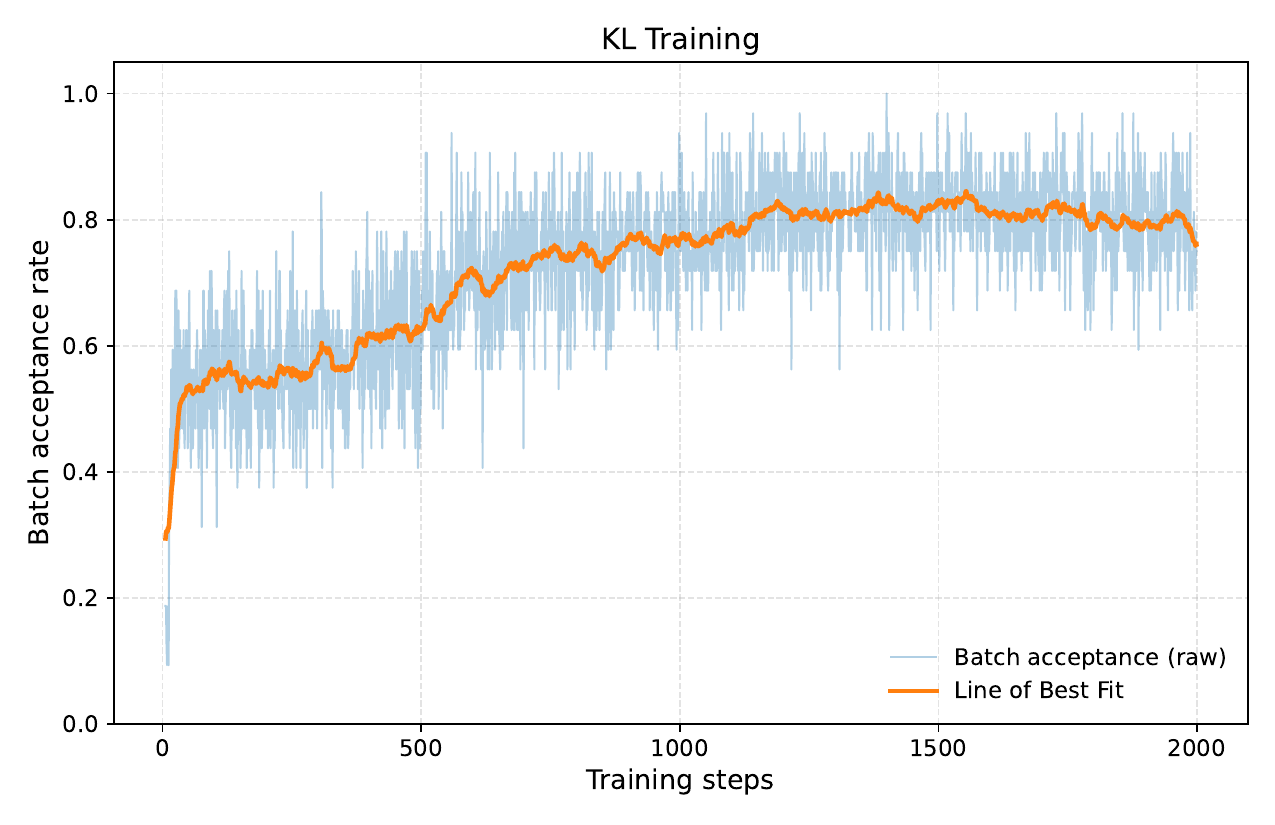}
    \subcaption{\textsc{KL-only}: stable, monotone gains.}
    \label{fig:abl-kl}
  \end{subfigure}
  \hfill
  \begin{subfigure}[h]{0.32\textwidth}
    \centering
    \includegraphics[width=\linewidth]{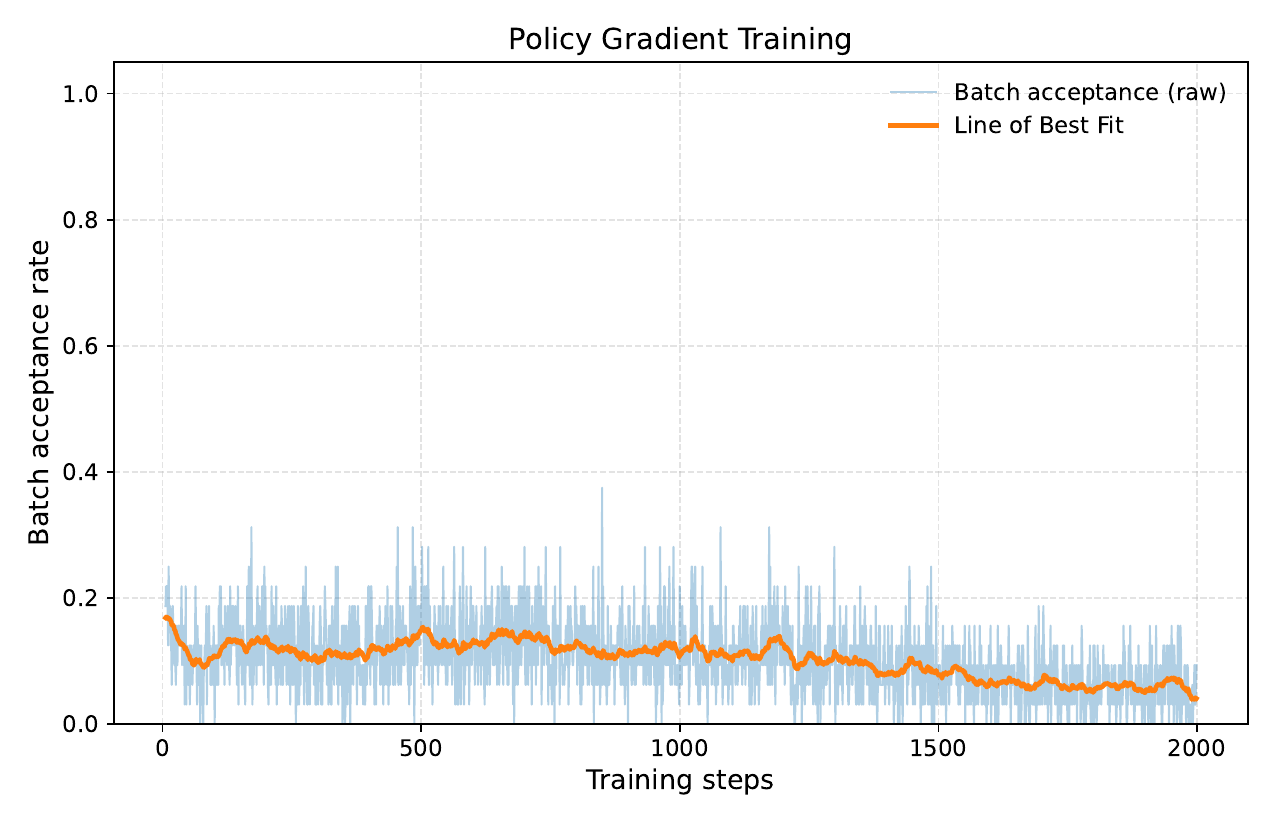}
    \subcaption{\textsc{PG-only}: flat, noisy learning curve.}
    \label{fig:abl-pg}
  \end{subfigure}
  \hfill
  \begin{subfigure}[h]{0.32\textwidth}
    \centering
    \includegraphics[width=\linewidth]{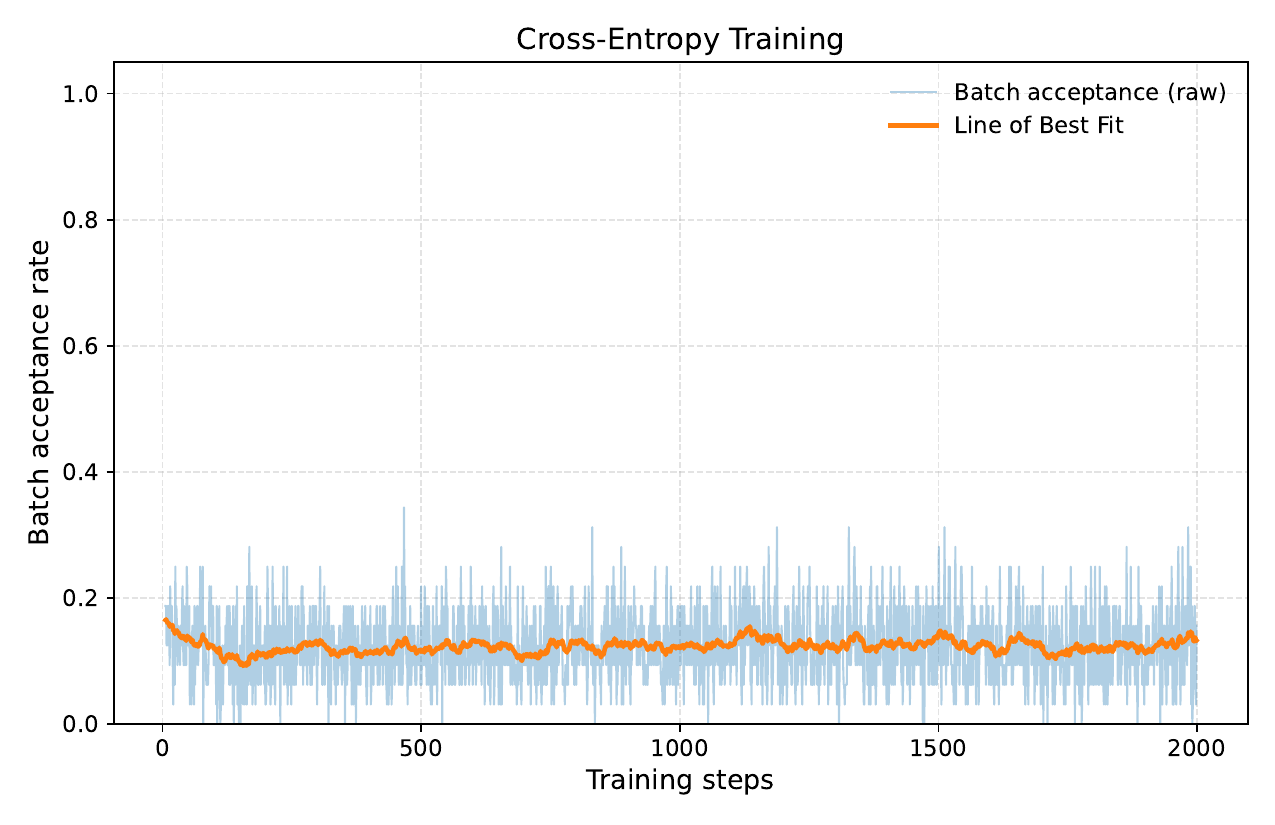}
    \subcaption{\textsc{CE-only}: flat learning curve.}
    \label{fig:abl-ce}
  \end{subfigure}
  \vspace{-2mm}
  \caption{\textbf{Objective ablations:} Batch acceptance rate vs.\ training steps.
  Curves computed on the same data stream, split, and $k_{\text{spec}}$ as the main setup.}
  \label{fig:ablation-acceptance-triptych}
  \vspace{-3mm}
\end{figure*}

\begin{table}[h]
\centering
\caption{Objective ablations on Spec-Bench (final). Higher is better.}
\label{tab:objective-ablation-final}
\begin{tabular}{lcc}
\toprule
Objective & Mean accepted tokens (MAT) & Speedup \\
\midrule
\textsc{KL-only} & 1.933 & \textbf{$1.435\times$} \\
\textsc{PG-only} & 0.035 & $0.341\times$ \\
\textsc{CE-only} & 0.039 & $0.335\times$ \\
\bottomrule
\end{tabular}
\end{table}

\paragraph{KL-Only:} We observe that online KD is sufficient to raise acceptance.
The batch-acceptance curve increases steadily and smoothly across training (Fig.~\ref{fig:abl-kl}), indicating consistent improvements in draft/verifier agreement. However, the the acceptance curve begins to level out around a 80\% batch acceptance rate, indicating that there are limits to KL alone. 

This is further verified by the end results, as the KL-only training attains the highest speedup among single-term objectives (Table~\ref{tab:objective-ablation-final}), but still falls short of our full DVI pipeline. This demonstrates that because gradients are dense and low variance, KL alone can bootstrap a useful drafter, which empirically validates our KL-heavy warmup to avoid a cold start in RL.

\paragraph{PG-Only:} We observe that sparse rewards and censored feedback hinder learning. Acceptance remains near zero and exhibits noise rather than an upward trend (Fig.~\ref{fig:abl-pg}). The end-state MAT and speedup indicate the model is $\sim3\times$ slower than a baseline model (Table~\ref{tab:objective-ablation-final}).

With a frozen verifier and shallow draft, rewards are extremely sparse as only agreement with the verifier provides $r{=}1$. This “bandit with censoring” produces high-variance gradients and weak credit assignment. Without KD to keep logits calibrated, exploration drifts, further reducing agreement and compounding variance. These results support keeping PG as a light, on-policy correction after KD stabilization.

\paragraph{CE-only:} We observe that reward-masked supervision is too weak without calibration. Similar to the PG-only objective, acceptance stays flat over training (Fig.~\ref{fig:abl-ce}), and end-state MAT and speedup indicate a model $\sim3\times$ slower than the baseline model (Table~\ref{tab:objective-ablation-final}).
Because only accepted tokens are labeled, the CE target distribution is censored and narrow. Similar to PG-only, the distribution-level guidance of KL is necessary to improve acceptance.

\paragraph{Insights}
The ablations confirm that 
\begin{enumerate}
    \item KL-only reliably increases acceptance (dense, low-variance signal).
    \item PG-only and CE-only struggle under sparse/censored feedback.
\end{enumerate}

It should be note that many sampling-based SD methods and Online SD methods rely almost exclusively on KL / distillation for teaching the drafter \citep{liu2024onlinespeculativedecoding, zhou2023distillspec}. We demonstrate that in a greedy decoding setting, strict distillation is significantly slower than DVI's approach. 

These results empirically justify DVI’s staged objective as described in Section~\ref{sec:method}. The schedule avoids RL cold start, preserves losslessness under the same sampler, and keeps training cheap and robust.

\section{Conclusion}\label{conclusion}

We presented \emph{Draft, Verify, \& Improve (DVI)}, a training-aware self-speculation method that closes the loop between inference and learning. DVI splits a single backbone into shallow drafting and deep verifying paths, keeps the verifier frozen, and updates a lightweight LoRA drafter online from accept/reject feedback.

When evaluated on Spec-Bench, DVI delivers $2.16\times$ end-to-end speedups on average, matching SoTA methods like EAGLE-2, and surpassing them on Translation, QA, and RAG. 
DVI attains these gains with a dramatically smaller training budget: 2{,}000 online prompts versus the \textit{millions} of prompt exposures used by other methods. 

Ablation studies validate our training pipeline. Online KL (distillation) alone produces steady acceptance gains, but plateaus; PG-only and reward-masked CE-only objectives struggle due to sparse/censored rewards and poor calibration. Our proposed KL$\rightarrow$RL schedule stabilizes early learning and adds targeted on-policy improvements, yielding the full DVI performance.

\bibliography{iclr2026_conference}
\bibliographystyle{iclr2026_conference}

\appendix

\section{Experimental Setup and Reproducibility}
\label{app:setup}

\paragraph{Hardware.}
All experiments were run on a single NVIDIA H100 GPU. To check hardware sensitivity, we repeated a subset of runs on an NVIDIA A40; qualitative conclusions and relative rankings were unchanged. Due to compute and time constraints, we did not replicate all experiments on different GPUs. 

\paragraph{Model and decoding configuration.}
All results use the Vicuna-7B backbone specified by \textsc{Spec-Bench}. We adopt the self-speculative split from the main text: the draft path comprises the shallow layers up to index $k{=}2$ and the target path comprises the remaining layers ($3{\to}L$). The drafter head is LoRA-parameterized and trainable; the verifier head and backbone are frozen. Unless otherwise stated, we use greedy decoding (temperature $0$) for verification, proposal depth $k_{\text{spec}}{=}4$, and the tokenizer and context limits mandated by the benchmark.

\paragraph{Evaluation protocol.}
We evaluate through the \textsc{Spec-Bench} harness, which standardizes tokenization, decoding policy, batching, and logging across methods by testing them in the same environment and hardware. Competing methods are invoked via the official public implementations referenced by \textsc{Spec-Bench} (e.g., GitHub/Hugging Face integrations). 

We do not retrain external baselines; when a method exposes a user knob (e.g., draft depth), we use the authors’ recommended defaults as surfaced by the harness. Metrics follow the benchmark: mean accepted tokens (MAT) and end-to-end wall-time speedup relative to greedy autoregressive decoding of the baseline model.

\paragraph{Absolute vs.\ relative speedups.}
Despite using the official public implementations within a common harness, we observe absolute speedups that are lower than the numbers reported in the respective papers for all methods. Replicating on an NVIDIA A40 produced similar absolute speedups to the H100 in our setup, e.g. lower than reported in respective papers. 

Importantly, the \emph{relative} ordering is consistent with prior reports: EAGLE-family methods are fastest, followed by Hydra, then Medusa, etc. Our conclusions therefore focus on the apples-to-apples comparisons within the shared \textsc{Spec-Bench} environment and the relative efficiency of DVI under identical settings.

\paragraph{Reproducibility}
We will release code, configuration files (including split index, $k_{\text{spec}}$, and decoding settings), and scripts to reproduce all tables and figures upon publication.

\section{AI Acknowledgment}
We used AI-based assistants for stylistic polishing and editing language. All technical ideas, analyses, conclusions, etc. are our own.

\end{document}